\documentclass{article}
\usepackage{spconf,amsmath,epsfig}
\usepackage{stackengine} 
\newcommand\oast{\stackMath\mathbin{\stackinset{c}{0ex}{c}{0ex}{\ast}{\bigcirc}}}

\usepackage{cite}

\usepackage{graphicx}
\graphicspath{{Figures/}}
\DeclareGraphicsExtensions{.pdf,.jpeg,.png}
\usepackage{amsmath}
\usepackage{algorithmic}

\usepackage{gensymb}
\usepackage{subfloat}

\usepackage{array}
\usepackage{multirow}
\usepackage{amsfonts}
\usepackage{stfloats}
\usepackage{url}
\usepackage[export]{adjustbox}
\usepackage{textcomp}
\usepackage{cite}

\usepackage{caption}
\usepackage[font=footnotesize]{subcaption}
\usepackage{breqn}
\begin{document}
	%
	\title{Object counting from aerial remote sensing images: \\ application to wildlife and marine mammals}
	
	
	%
	\name{Tanya Singh $^*$, Hugo Gangloff $^{*,\dagger}$, Minh-Tan Pham $^*$  \thanks{This work was done in the context of the SEMMACAPE project, which benefits from an ADEME ({\em Agence de la transition écologique}) grant under the “Sustainable Energies” call for research projects (2018--2019).}}

	\address{ $^*$ IRISA, Université Bretagne Sud, UMR 6074, 56000 Vannes, France \\
		$^{\dagger}$ INRAE, AgroParisTech, UMR MIA Paris-Saclay, 91120 Palaiseau, France\\
		\texttt{\ninept singh.tanya3298@gmail.com,minh-tan.pham@irisa.fr,hugo.gangloff@inrae.fr}}


	\maketitle
	

	\begin{abstract}

        Anthropogenic activities pose threats to wildlife and marine fauna, prompting the need for efficient animal counting methods. This research study utilizes deep learning techniques to automate counting tasks. Inspired by previous studies on crowd and animal counting, a UNet model with various backbones is implemented, which uses Gaussian density maps for training, bypassing the need of training a detector. The new model is applied to the task of counting dolphins and elephants in aerial images. Quantitative evaluation shows promising results, with the EfficientNet-B5 backbone achieving the best performance for African elephants and the ResNet18 backbone for dolphins. The model accurately locates animals despite complex image background conditions. By leveraging artificial intelligence, this research contributes to wildlife conservation efforts and enhances coexistence between humans and wildlife through efficient object counting without detection from aerial remote sensing.

        	\end{abstract}


\section{Introduction}
\label{sec:1-Introduction}

To study and control the harmful effects of anthropogenic activities on wildlife and marine fauna, researchers depend heavily on the count of animals \cite{matthiopoulos2020species}. 
Ecologists and biologists conduct surveys to get the species count, monitor their population and track their migratory patterns. This information can help understand the health and stability of ecosystems and can be used to monitor conservation and management efforts. Historically, the counting for these ecological surveys was done manually. However, with the advent and progress of artificial intelligence, researchers have been looking for ways to automate the counting task by leveraging the potential of machine learning and then, recently, deep learning. 

Deep learning techniques can make this long and tedious task faster and more efficient, by allowing the processing of large volumes of data and delivering results devoid of human errors \cite{christin2019applications}.
Object counting from images is among the popular vision tasks, which can be done by using detection-based approach or regression-based methods \cite{fan2022survey}. For the first approach, standard object detection frameworks (which predicts bounding boxes around objects) can be applied, then the number of objects is counted based on the predicted boxes. Not necessarily requiring bounding box labels but point annotation, the second counting approach can be performed based on regression network, usually relying on density maps (indirect counting) or not (direct counting). In this research work, inspired by the crowd counting studies in computer vision \cite{wan2019adaptive} and building up from works making use of density maps \cite{padubidri2021counting}, we develop a UNet-based model for counting marine mammals (e.g. dolphin) at offshore wind farms, as well as for counting wildlife elephants from aerial remote sensing images. Our experiments conducted on two dedicated datasets show the high potential of our approach for this task. 

In the following section, we present the two studied datasets. Sec.~\ref{sec:2-Methodology} then describes the developed method and its implementation details. We provide quantitative and qualitative results in Sec.~\ref{sec:3-ExpResults} before drawing conclusions in Sec.~\ref{sec:4-Conclusions}.

\section{Datasets}
Two aerial image datasets are studied in this work. The first one is the Aerial Elephant Dataset (AED) which was proposed by \cite{naudé2019aerial} to promote research on animal detection in practical conditions. This dataset consists of 2101 aerial images with dot annotations for a total of 15,511 African bush elephants in their natural habitat, with complex backgrounds, as seen in Fig.~\ref{fig:ground_truth}.
The second one is the Semmacape Dataset (SD) \cite{berg2022weakly, gangloff2022variational} which contains 165 box-annotated aerial images collected in the Gironde estuary and Pertuis sea Marine Nature Park, France, during spring 2020. The background of this dataset is quite complex due to the appearance of sun glare and waves which act as noise within the images. Also, the marine animals are at different distance to the sea surface and can have various shapes. Recognizing those under-water animals become quite challenging (see Fig.~\ref{fig:ground_truth}). 



\section{Methodology}
\label{sec:2-Methodology}
The approach we develop aims at counting animals without detection. The advantage of such regression-based methods are that they avoid training and learning a detector to predict object coordinates and directly target the counting task \cite{chen2012feature}. Moreover, they are less costly in annotations (dots or scalar numbers are required as inputs, not bounding box annotations). In \cite{lempitsky2010learning}, the authors considered pioneers for object counting through density map estimation using dot annotations. They proposed a supervised learning framework to generate pixel-level ground-truth density maps from dot annotations using Gaussian kernels. The count can be obtained by integrating over the Gaussian density map. The density map approach is faster than detection-based counting methods and preserves spatial information, as opposed to the direct regression-based counting strategy \cite{fan2022survey}. 

\subsection{Ground-truth density maps}
For both datasets, AED\cite{naudé2019aerial} and SD \cite{berg2022weakly}, the ground-truth density maps are generated using dot-annotated images (which symbolize the centre of each object) convolved with a Gaussian filter with a fixed variance ($\sigma^2$). The variance of the Gaussian filter is commonly based either on the size of the object of interest (for low density image) \cite{padubidri2021counting,hoekendijk2021counting} or the distance between the objects of interest (for high density images) \cite{aich2018improving}. We now describe these two techniques.

\subsubsection{Constant $\sigma^2$ -  Low density images}\label{sigma_low}
In Equation~\ref{eq:gaus}, the ground-truth density map, $DM_I$, for a dot-annotated image $I$, where $A_I$ represents the 2 coordinates of the dots, is defined as the sum of the Gaussian distributions centered at each dot-annotation in the image, evaluated at pixel intensity $p$ is computed as follows:
\begin{equation} \label{eq:gaus}
\begin{split}
DM_I (p) = \sum_{\mu \in A_I} \mathcal{N}(p ; \mu,\,\Sigma),
\end{split}
\end{equation}
where $\mathcal{N}(p ; \mu,\,\Sigma)$ represents a Gaussian distribution with mean $\mu$ and covariance matrix $\Sigma$; $\Sigma=\mathrm{diag}(\sigma^2)$.

\begin{figure}[h!]
\centering
\includegraphics[width=0.4\textwidth]{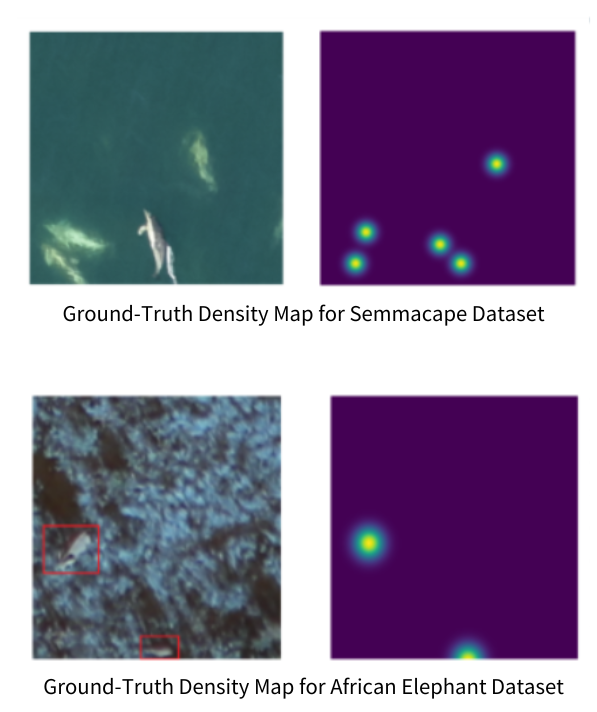}
\caption{Illustrations of the images (left) and the ground-truth density maps (right) from the two studied datasets.}
\label{fig:ground_truth}
\end{figure}

Fig.~\ref{fig:ground_truth} shows the generated ground-truth density maps for the two datasets, using this method.

\subsubsection{Adaptive $\sigma^2$ -  High density images}\label{sigma_high}
In the case of high-density images (e.g., for crowd counting), two objects close by can have overlapping distributions if the $\sigma^2$, is not changed. Therefore, it is necessary to compute for each object $(x_i)$ in the image the average distance $d_{\mathrm{avg},i}$ to its $k$-nearest neighbour:
\begin{equation}\label{eq:GAM3}
\begin{split}
d_{\mathrm{avg},i} = \frac{1}{k} \sum_{j=1}^{k} d_{ij},
\end{split}
\end{equation}
where $d_{i,j}$ is the distance between two objects $i$ and $j$.
Then, to get the density around each point $(x_i)$, the image has to be convolved with a Gaussian kernel with variance $\sigma_i^2$ given by the product of $d_{\mathrm{avg},i}$  and $\sigma_0^2$, the initial variance as suggested in \cite{aich2018improving,li2018csrnet}. The computation of this density map version is given by:
\begin{equation}\label{eq:GAM4}
\begin{split}
DM_I = H(x) \oast P_{\sigma_i}(x) \text{ where } \sigma^2_i = \sigma_0^2 d_{\mathrm{avg},i},
\end{split}
\end{equation}
where $\oast$ denotes the convolution operator.

\subsubsection{Count from density maps}
The count of objects $N_I$ from a density map $DM_I$ can be obtained by integrating the pixel values in the density map:
\begin{equation} \label{eq:count_den}
\begin{split}
N_I = \sum_{p \in I} DM_I (p).
\end{split}
\end{equation}

\subsection{Model implementation}
\begin{figure}[h!]
\centering
\includegraphics[width=0.5\textwidth]{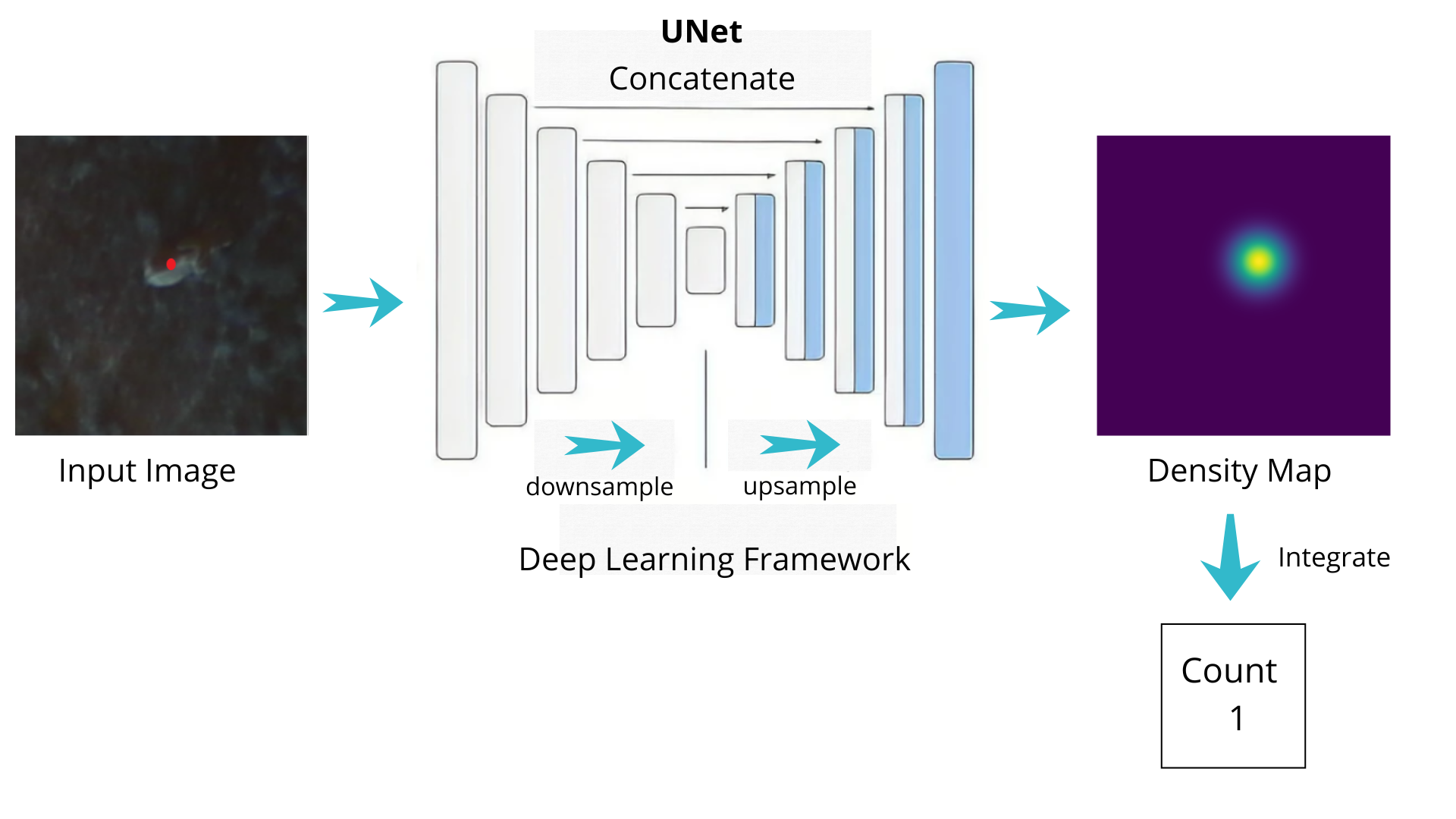}
\caption{The proposed method (inspired by \cite{padubidri2021counting}) for animal counting. The image is fed into the UNet architecture to produce a density map, the integration of which gives the count.}
\label{fig:model_unet}
\end{figure}

We train our UNet model using the generated ground-truth density maps of constant $\sigma^2$ (as animals in the two studied datasets are quite sparse). The UNet architecture preserves high- and low-level features, which enables the accurate object identification and localization in images \cite{he2016deep}. The UNet encoder creates a high-dimensional feature vector, while the decoder generates the density map by projecting features to the pixel space \cite{he2016deep}. The count can be then obtained by integrating over the produced density map. The following illustration in Fig.~\ref{fig:model_unet} shows the proposed approach. 

For the Unet backbone, we explore both the ResNet and EfficientNet, pretrained on ImageNet. For the AED, 2653 were used for training, 460 for validation and 460 for the test. For the SD, we used 91 images for training, 31 for validation and 31 for test. 

\subsection{Loss function}
The Root-Mean-Square Error (RMSE) between the predicted density map ($\widehat{DM}$) and the original density map ($DM$) is used as the objective function of our model. It is given by:
\begin{equation} \label{eq4}
\begin{split}
RMSE = \sqrt{\frac{1}{N} \sum_{n=1}^{N} (\widehat{DM}-DM)^2}.
\end{split}
\end{equation}

\section{Experimental results}
\label{sec:3-ExpResults}

\subsection{Quantitative results}
To measure the performance of our method, two metrics are used which are the RMSE and the MAE (Mean Absolute Error), computed by the following equations.

\begin{equation} \label{eq:rmse}
\begin{split}
RMSE = \sqrt{\frac{1}{N} \sum_{i=1}^{N} (\hat{y}_i-y_i)^2},
\end{split}
\end{equation}
where $\hat{y}_i$ and $y$ are respectively the predicted count and the real count for the $i^{th}$ image.

\begin{equation} \label{eq:mae}
\begin{split}
MAE = {\frac{1}{N} \sum_{i=1}^{N} |\hat{y}_i-y_i|}.
\end{split}
\end{equation}



Table \ref{tb:results} provides the performance of our model on the two studied datasets.
Our implementation of the UNet model with EfficientNet-B5 backbone resulted in better RMSE and MAE values (0.89 and 0.49, respectively) for the African Elephant dataset, which is consistent with the results obtained by Padubidri et al., \cite{padubidri2021counting}. Notably, our model outperformed the count-ception approach \cite{bar2018biomass} which provided an RMSE value of 1.59. Similarly, we got the RMSE and MAE values of 0.713 and 0.35 for dolphin counting in the Semmacape dataset, respectively, by using the UNet model with ResNet18 backbone. 

\begin{table}[htb]
\centering
\begin{tabular}{ |c|c|c|c| } 
 \hline
 Dataset & Method & RMSE & MAE \\ 
 \hline
AED & UNet (EfficientNet-B5) & 0.890 & 0.490 \\ 
 \hline
SD & UNet (Resnet18) & 0.713 & 0.350 \\ 
 \hline
\end{tabular}
\caption{The quantitative results (RMSE and MAE) of the developed model for animal counting from the two studied datasets.}
\label{tb:results}
\end{table}

\subsection{Qualitative results}
\begin{figure*}[htb]
    \centering
    \includegraphics[width=0.99\textwidth]{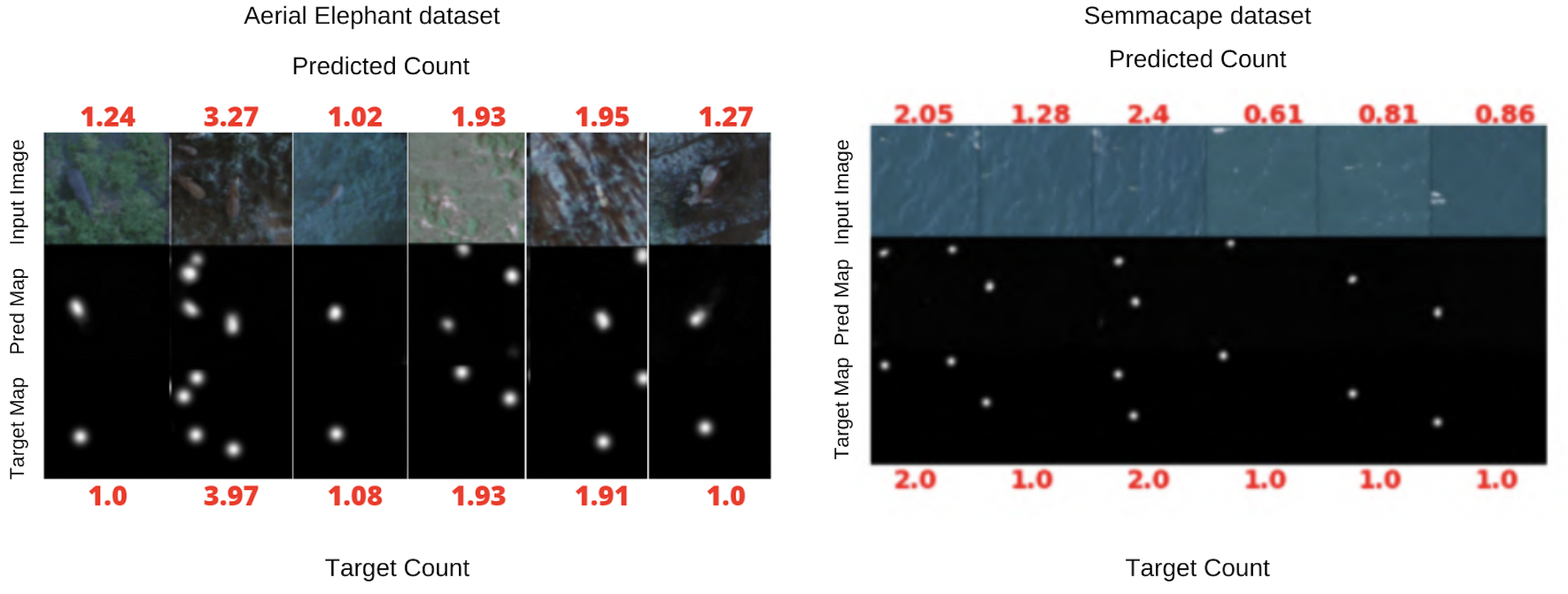}
    \caption{Visualization of the output density maps from the model on two datasets. The top row with red numbers is the predicted count from the model; the bottom row is the ground truth count. The top image row is the input image; the middle is the predicted density map; the bottom image is the ground truth density map.}
    \label{fig:results}
\end{figure*}

We provide some visual results yielded by the developed network in Fig.~\ref{fig:results}. As observed, our model performs quite well in locating the elephants (in AED images on the left) and dolphins (in SD images on the right) regardless of the complex backgrounds with different illumination, dolphins under the surface, sun glare, waves and occlusion of the animals. 

There is certainly scopes for improvement in counting results on the studied datasets. For instance, pre-processing can be adopted to remove noise like sun-glare and waves from Semmacape data, or to enhance the image contrast from the Elephant dataset. We observe that these problems of noise and low contrast are the main factors to cause erroneous counts.

\section{Conclusion}
\label{sec:4-Conclusions}

The objective of this research work is to count animals from aerial imagery without detection using regression-based approach with Gaussian Density Maps. Counting without detection is important because in many cases, detecting every object instance is not feasible and necessary. We have implemented a simple and efficient method that is able to count objects in aerial imagery. We have applied and showed the potential of the method to count elephants and dolphins in both wildlife and marine scenarios. 

One of the perspective studies could be to extend the model to count the different species of animals (e.g., for the Semmacape dataset). By this way, the class-wise count can be generated. Another potential direction to improve the model using self-supervised learning techniques \cite{berg2022self,liu2022leveraging} to integrate more powerful losses (i.e., contrastive) as well as to perform model pre-training in case of a large amount of unlabeled data are available.


\bibliographystyle{ieeetr}
\bibliography{refs}
\end{document}